# Active Use of Latent Constituency Representation in both Humans and Large Language Models

Wei Liu[a], Ming Xiang[b], Nai Ding[a, c]*

[a] Key Laboratory for Biomedical Engineering of Ministry of Education,
College of Biomedical Engineering and Instrument Sciences,
Zhejiang University, Hangzhou, China

[b] Department of Linguistics,
The University of Chicago, Chicago, USA

[c] State Key Lab of Brain-Machine Intelligence; MOE Frontier Science Center
for Brain Science & Brain-machine Integration,
Zhejiang University, Hangzhou, China

*Corresponding author：

Nai Ding

College of Biomedical Engineering and Instrument Science,

Zhejiang University, Hangzhou, China

E-mail address: ding_nai@zju.edu.cn

**Abstract**

Understanding how sentences are internally represented in the human brain, as well as in large language models (LLMs) such as ChatGPT, is a major challenge for cognitive science. Classic linguistic theories propose that the brain represents a sentence by parsing it into hierarchically organized constituents. In contrast, LLMs do not explicitly parse linguistic constituents and their latent representations remains poorly explained. Here, we demonstrate that humans and LLMs construct similar latent representations of hierarchical linguistic constituents by analyzing their behaviors during a novel one-shot learning task, in which they infer which words should be deleted from a sentence. Both humans and LLMs tend to delete a constituent, instead of a nonconstituent word string. In contrast, a naive sequence processing model that has access to word properties and ordinal positions does not show this property. Based on the word deletion behaviors, we can reconstruct the latent constituency tree representation of a sentence for both humans and LLMs. These results demonstrate that a latent tree-structured constituency representation can emerge in both the human brain and LLMs.

**Introduction**

Understanding how the human brain processes sentences is a central question in multiple disciplines such as linguistics, psychology, and cognitive science. A hypothesis, which was revolutionary when first proposed and remains highly influential until today, is that the essential operation of sentence comprehension is to convert the sentence from a linear sequence of words into hierarchically organized constituents (Bloomfield, 1933; Chomsky, 1957). Many findings in psychology and neuroscience support this hypothesis (Hauser et al., 2002; Friederici et al., 2011, 2017). For example, non-human animals have trouble learning human language (Fitch et al., 2005) and a likely reason is that animals do not have the capacity to parse hierarchically organized constituents – animal communication signals lack constituency structures and animals fail on tasks that require recursion, a core operation in constituency analysis (Fitch & Hauser, 2004). Furthermore, many neuroimaging and neurophysiology experiments show that the neural response to sentences is better explained by hierarchical linguistic constituents than merely linear word sequence (Pallier et al., 2011; Brennan et al., 2016; Ding et al., 2016; Nelson et al., 2017, see also Frank et al., 2012).

With the success of large language models (LLMs) in many language processing tasks (Zeng et al., 2023; Touvron et al., 2023; OpenAI et al., 2024), a new question has arisen regarding to how LLMs encode sentences. Unlike the human brain, the current LLMs are deep neural networks (DNNs) whose network connections are artificially designed and well understood

(Vaswani et al., 2017; Brown et al., 2020). Nevertheless, the high complexity of the network connections makes it difficult to explain how these connections generate language behavior, which is referred to as the explainability problem (Marcus & Davis, 2019; Arrieta et al., 2020; Hagendorff, 2023) and echoes Marr's division between implementation, algorithm, and computation (Marr, 1982). The LLMs are not designed to encode the hierarchical constituency structure of a sentence, but linguistic constituents are abstract constructs that still have the potential to functionally explain the behavior of LLMs – for example, it has been shown that the internal tree-structured representation of a sentence could be reconstructed based on responses of a deep neural network that is not trained on any constituency analysis task (Hewitt & Manning, 2019; Kim et al., 2019; Arps et al., 2022). It remains unclear, however, whether the tree-structured representation actively contributes to language comprehension or is just a redundant representation (Belinkov, 2022; van Dijk et al., 2023).

The lack of empirical methods that can reveal the potential constituency representation of a sentence during active language use is a major obstacle towards understanding latent sentence representation in both the human brain and LLMs. Developing such methods, however, faces two challenges. First, both humans and LLMs can do constituency parsing after training (Tian et al., 2020; Zhang et al., 2020), but this is not evidence that humans or LLMs automatically apply constituency analysis during natural language comprehension. Therefore, the potential latent constituency representation must be implicitly tested. Second, since LLMs are trained on a huge amount of text, it is challenging to find testing materials that do not overlap with their training samples. If the testing materials overlap with the training samples, it cannot be ruled out that the LLMs solve the task utilizing shallow statistical cues (Bender & Koller, 2020; Bender et al., 2021; Binz & Schulz, 2023; Wilson et al., 2023).

Here, we aim to develop a method that can reveal the potential constituency representation of a sentence by analyzing language-use behavior. The method is based on a novel one-shot learning task, in which the participants learn to delete a word string from a sentence based on a single demonstration (Fig. 1a). The task does not require any explicit linguistic knowledge and is unfamiliar to both humans and LLMs. In the task, we always deleted a constituent in the demonstration but, with just one demonstration, the underlying deletion rule was ambiguous, and the participant might infer the rule based on, e.g., the semantic or syntactic properties of individual words, the ordinal position of words within a sentence, or the syntactic category of a linguistic constituent. In this unfamiliar but easy-to-follow task, we investigated (1) whether humans and LLMs preferred to delete a constituent, (2) what rules were inferred by humans and LLMs in the task, (3) whether a full constituency tree could be reconstructed based on the word deletion task and how similar this constituency tree compared with the linguistic constituency tree.

# Probe Sensitivity to Constituency Through a Word Deletion Task

**a** Experiment procedure

demonstration (1-shot):
*She had an idea* ➡ *She had ~~an idea~~*

test sentence:
*We put some order together* ➡ ?

| human/ChatGPT response | data analysis |
|---|---|
| test 1<br>*We put together* | constituent<br>*~~some order~~* |
| ⋮ | ⋮ |
| test N<br>*We put some* | nonconstituent<br>*~~order together~~* |

**b** Demonstration in Experiment 1: delete an NP embedded in a VP

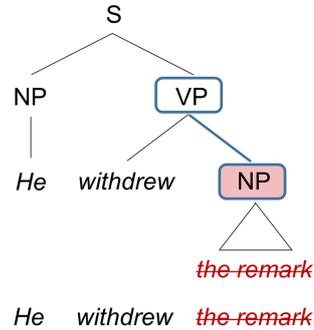

**c** Properties of deleted word string in Experiment 1

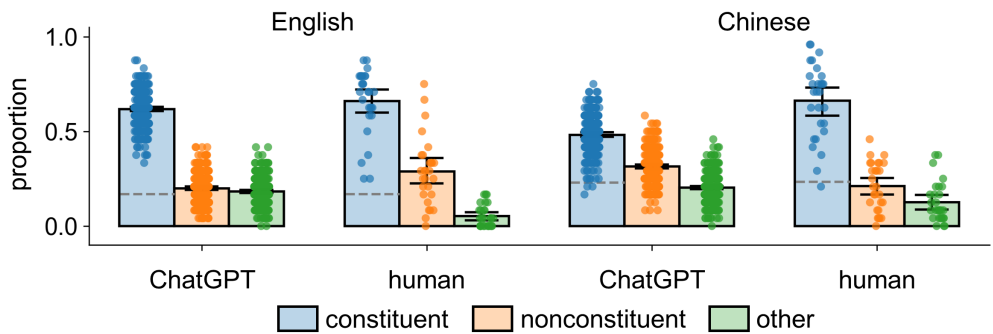

constituent — nonconstituent — other

**d** Constituent rate of LSTM using few-shot learning in Experiment 1

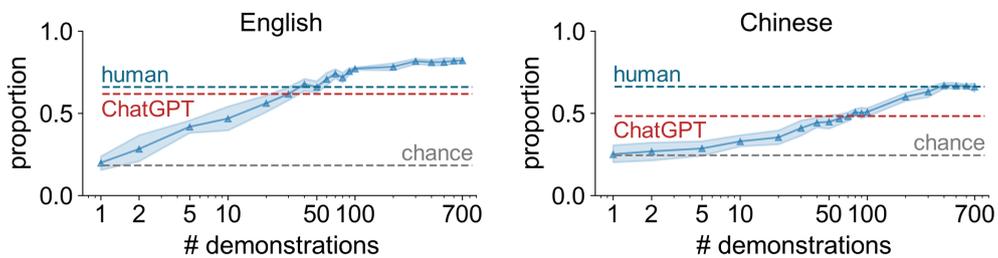

Figure 1. The word deletion task and constituent rate in Experiment 1. **a.** In each test, the participants delete a word string from a test sentence following a single demonstration. **b.** Demonstration in Experiment 1, in which we delete an NP that is embedded in a VP. **c.** Properties of the deleted word string, i.e., whether it is a linguistic constituent. The constituent rate is significantly higher than the chance level denoted by the dashed gray line. The label 'other' indicates invalid responses. Error bars represent 95% confidence intervals (CIs) of the mean across participants (for humans) or runs (for ChatGPT), estimated using bootstrap. **d.** Constituent rate of a naive LSTM model that is trained to perform the word deletion task based on different numbers of demonstrations. The model has access to word properties and word position as the input, but is not pretrained on human language. Shaded areas cover

95% CIs of the mean proportion across runs, estimated using bootstrap. Mean performance of humans and ChatGPT in the one-short learning experiments is shown by the dashed lines.

**Results**

In a novel word deletion experiment, humans or LLMs were given a single demonstration in which a word string was deleted from a sentence. They had to infer the underlying rule and apply it to a test sentence. In each test, a single demonstration was paired with a single test sentence. The word string deleted in the demonstration was always a complete linguistic constituent – in a constituency tree representation (e.g., Fig. 1b), words in the sentence were grouped into latent units, e.g., noun phrases (NPs) and verb phrases (VPs), and these latent units are referred to as constituents. In the experiment, however, we did not explicitly inform the participants, whether a human participant or a LLM, anything about linguistic constituents. For the example in Fig. 1a, the full instruction to humans, also the prompt to LLMs, was "*John developed a very special way of speaking. For 'She had an idea', he would say 'she had'. Please carefully examine John's speaking style, and guess what he would say for the sentence: 'We put some orders together'*". Note that, although the task was referred to as a word deletion task in this paper, the participants were not explicitly told that they had to delete words during the experiment.

We recorded the response from humans or LLMs and took the difference between the recorded word string and the original sentence to obtain the deleted word string. For example, in Fig. 1a, if the participant typed in "*We put together*", the deleted word string was "*some order*", which was a constituent. In contrast, if the participant typed in "*We put some*", the deleted word string was, i.e., "*order together*", which was not a constituent. In the following, we analyzed whether human participants and LLMs preferred to delete a word string that formed a complete linguistic constituent. In most of the experiments, we tested ChatGPT (OpenAI, 2022) as a representative LLM, and for the most general experiment, i.e., Experiment 2, we further tested four other LLMs, i.e., GPT-4, Claude-3 , Gemini-1, and Llama-3.

**Sensitivity to Linguistic Constituency**

Experiment 1 was presented in either Chinese or English, to either human participants or ChatGPT. In the experiment, both the demonstration sentence and the test sentence were required to have an NP that was embedded in a VP. Furthermore, the NP was required to be a direct child node of VP in the demonstration, and a descendant node of VP in the test sentence (see Methods for the definitions of child and descendant nodes). In the demonstration, the NP embedded in a VP was deleted (Fig. 1b) and the participants were asked to infer the transformation and apply it to a test sentence.

We selected demonstration sentences ($N$ = 807 for English and $N$ = 716 for

Chinese) and test sentences ($N = 942$ for English and $N = 837$ for Chinese) with the aforementioned properties from the Penn Treebank (Marcus et al., 1993) and the Chinese Treebank (Xue et al., 2005), which contained syntactic annotations of articles from newspapers and magazines. To sample as many sentences as possible, we randomly selected 24 pairs of demonstrations/test sentences from the sentence pools for each human participant (or each run of ChatGPT). In the human experiment, participants were native speakers of the presented language (30 participants per language, 24 tests per participant). In the ChatGPT experiment, ChatGPT automatically detected the input language and always responded using the same language in our experiment (300 runs per language, 24 tests per run).

We first analyzed how frequently the deleted word string was a complete constituent using the constituent rate, i.e., the total number of deleted word strings that formed a complete linguistic constituent divided by the total number of tests ($N = 720$ for humans and $N = 7200$ for ChatGPT). The constituent rate was above the chance level estimated based on randomly selected word strings (see Methods), for both human participants (Fig. 1c, $p < 0.001$ for both Chinese and English, one-sided Monte Carlo test, false discovery rate (FDR) corrected) and ChatGPT ($p < 0.001$ for both Chinese and English, one-sided Monte Carlo test, FDR corrected). For tests in which the participants did not delete a complete linguistic constituent, the participants either deleted a nonconstituent word string (see nonconstituent rate in Fig. 1c) or failed to learn that the task suggested them to delete words, instead of adding words not in the original sentence or not changing the original sentence (labeled as 'other' in Fig. 1c).

For ChatGPT, each test was processed independently, and the human participants were also instructed that all tests were independent. Nevertheless, it remained possible that the participants adapted their rule as they were exposed to more tests, i.e., learning based on multiple demonstrations. Therefore, we also analyzed the response to each of the 24 tests that the human participants accomplished (Fig. S1). The constituent rate was consistently above chance ($p$ = 0.025 and 0.002 for the tests #3 and #12 in Chinese, $p$ < 0.001 for all other comparisons, one-sided Monte Carlo test, FDR corrected) and did not significantly vary across tests (English: $F(23, 621)$ = 0.881, $p$ = 0.626; Chinese: $F(23, 621)$ = 1.24, $p$ = 0.206. One-way repeated measures ANOVA).

To provide a stronger baseline that could infer rules based on word properties and the ordinal position of a word within a sentence, we trained a naive LSTM model to predict the deleted word string. The LSTM was a powerful neural network to discover statistical patterns in sequences. Here, the LSTM had access to the properties of individual words through the pretrained word embeddings and the ordinal position of a word within a sentence through position embeddings. Unlike ChatGPT that was trained on a large amount of text (but not using the word deletion task), the LSTM was only trained on the word deletion task (without pretraining on large corpora). Based on a single demonstration, the constituent rate of LSTM was not much higher than chance (Fig. 1d, $p$ = 0.221 for English and $p$ = 0.380 for Chinese, one-sided Monte Carlo test, FDR corrected). After training with more demonstrations, the constituent rate of LSTM gradually increased and reached the constituent rate of humans and ChatGPT after about 50 demonstrations.

These results highlighted that both human participants and ChatGPT preferred to delete complete linguistic constituents in an unfamiliar task that did not explicitly encourage constituency analyses. Furthermore, this phenomenon could not be explained by statistical cues in word properties and ordinal positions, which were exploited by the LSTM model when learning the deletion task. The experiment was also tested on human participants and ChatGPT using other prompts, and different types of prompts led to highly similar results (Fig. S2 & S3).

**Sensitivity to Syntactic Category**

After establishing that both humans and ChatGPT were sensitive to linguistic constituency, we further investigated the word deletion rules inferred by the participants. We focused on two potential rules: The node-category rule deleted a constituent that had the same syntactic category as the constituent deleted in the demonstration. In contrast, the parent-category rule deleted a constituent whose parent node had the same syntactic category as the parent node of the constituent deleted in the demonstration.

In Experiment 1, the node-category rule always deleted an NP from the test sentence while the parent-category rule always deleted the child node of a VP (Fig. 2a). To distinguish these two rules, we selectively analyzed test sentences in which the NP was a descendant but not a direct child of the VP in the constituency tree (roughly 1/3 of the test sentences, see Methods). An example is shown in Fig. 2a, in which an NP ("*the future*") was a direct child of a PP ("*for the future*") that was a direct child of a VP. The node-category rule predicted that participants would delete the NP, i.e., "*the future*", while the parent-category rule predicted that participants would delete the direct child of the VP, i.e., "*for the future*". We analyzed whether the human and ChatGPT responses were better explained by the node-category rule or the parent-category rule. For English, the parent-category rule explained more responses for both humans and ChatGPT (Fig. 2b, $p < 0.001$ for both humans and ChatGPT, paired two-sided bootstrap, FDR corrected). For Chinese, however, the node-category rule explained slightly but significantly more responses for both humans and ChatGPT ($p < 0.001$ for both humans and ChatGPT, paired two-sided bootstrap, FDR corrected).

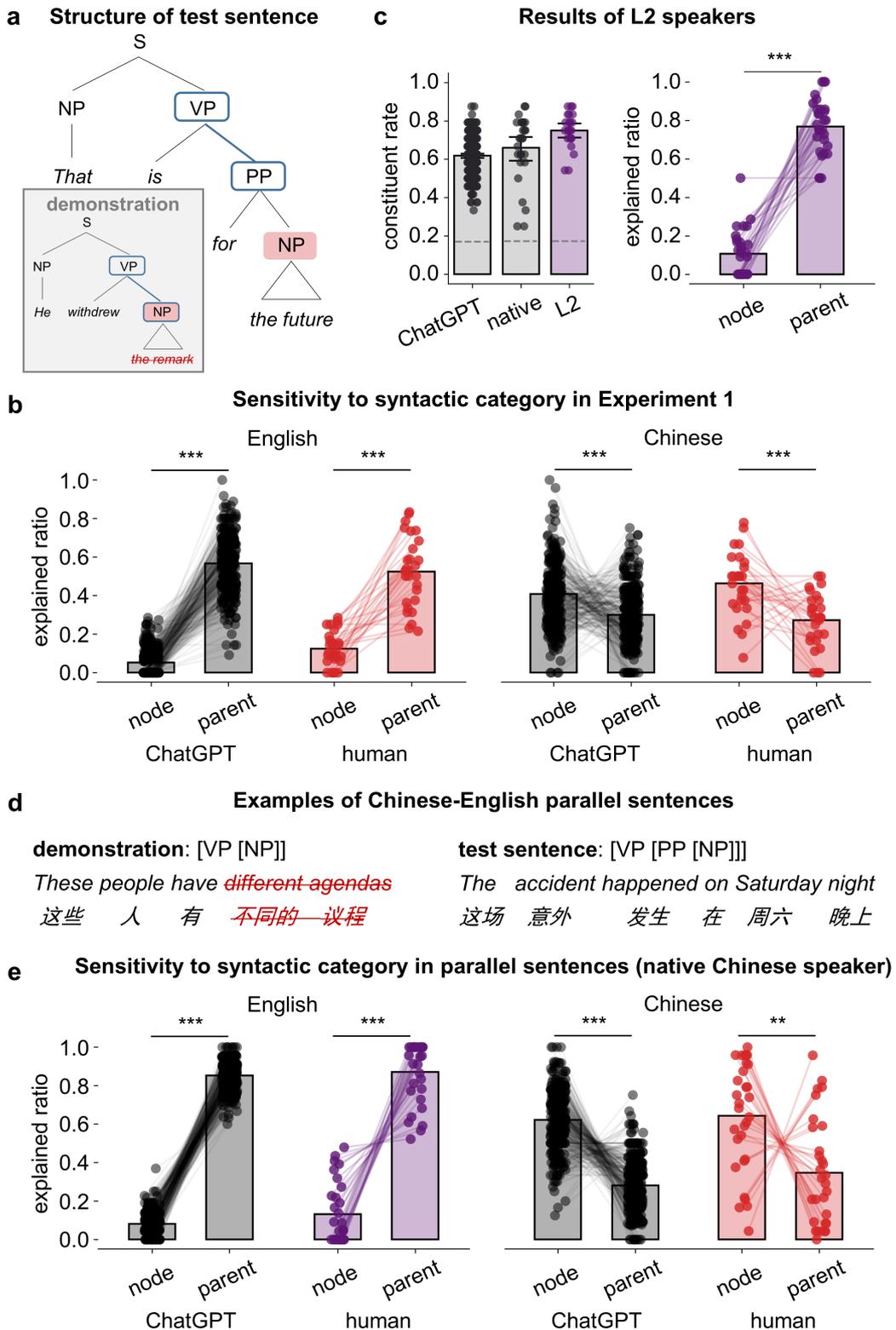

Figure 2. Sensitivity to syntactic category in Experiment 1 and influence of input language. **a.** Test sentence that can distinguish the parent-category rule and the node-category rule. Based on the demonstration in the inset figure (gray, same as Fig. 1b), the parent-category rule (blue border) deletes a child

node of VP, i.e., the PP, while the node-category rule (filled in red) deletes the NP. **b.** The explained ratio for the node- and parent-category rules. Points connected by a line represent a human participant or a single run of ChatGPT. **c.** Results of L2 speakers. *Left*: constituent rate of L2 speakers, in comparison with the results of native speakers and ChatGPT (gray, same in Fig. 1c). *Right*: explained ratio for each rule of L2 speakers. **d.** Parallel sentences that have the same syntactic structure and word order in Chinese and English. **e.** Explained ratio for the node- and parent-category rules for native Chinese speakers tested on either the Chinese (native language) or the English (second language) version of parallel sentences. **$p < 0.01$, ***$p < 0.001$.

Our results suggested both humans and ChatGPT inferred different rules when the same experiment was presented in different languages, i.e., Chinese and English. For humans, however, it remained unclear whether the preference of rules depended on the native language of a participant or if a participant could flexibly switch their preference depending on the input language. To distinguish these two possibilities, we recruited another group of native Chinese speakers who learned English as a second language (L2) to perform the English version of Experiment 1 (Fig. 2c). The results first confirmed that the L2 speakers also preferred to delete a constituent, and the constituent rate was even higher than that of native English speakers ($p = 0.011$, unpaired two-sided bootstrap, FDR corrected) and ChatGPT ($p < 0.001$, unpaired two-sided bootstrap, FDR corrected). Critically, L2 English speakers strongly preferred the parent-category rule ($p < 0.001$, paired two-sided bootstrap), similar to the preference of native English speakers, suggesting human participants could flexibly switch their preference for rule inference depending on the input language.

Why did participants prefer to infer different rules when analyzing different languages? A potential reason was that Chinese and English sentences tended to differ in their constituency structures. For example, modifiers could either precede or follow a noun in English but predominantly preceded a noun in Chinese. To test whether the preference of rules depended on the input language or the structure of the test sentences, we additionally constructed parallel sentences that shared the same constituency structure and word order in Chinese and English and tested these parallel sentences on both humans and ChatGPT (Fig. 2d). For both humans and ChatGPT, the constituent rate was significantly higher than chance (English: 0.93 ± 0.013 (95% CI) and 0.87 ± 0.0079 for humans and ChatGPT; Chinese: 0.97 ± 0.012 and 0.43 ± 0.011 for humans and ChatGPT), and the responses on English sentences were better explained by the parent-category rule (Fig. 2e, $p < 0.001$ for both humans and ChatGPT, paired two-sided bootstrap, FDR corrected) while the responses on Chinese sentences were better explained by the node-category rule ($p = 0.005$ for humans and $p < 0.001$ for ChatGPT, paired two-sided bootstrap, FDR corrected). These results demonstrated that, for both humans and ChatGPT, the preference for rule inference depended on the input language, instead of the specific sentences being tested.

**Generalization to Arbitrary Constituents**

Experiment 1 only tested one type of constituent, i.e., an NP embedded in a VP structure, and in Experiment 2 we tested whether the conclusions could generalize to an arbitrary constituent (Fig. 3). In Experiment 2, we randomly selected sentences from the Penn Treebank and the Chinese Treebank and deleted an arbitrary constituent to create demonstrations. In this experiment, the constituent rate of humans and ChatGPT remained significantly higher than chance (Fig. 3a, $p < 0.001$ for all comparisons, one-sided Monte Carlo test, FDR corrected), and the constituent rate was similar to the corresponding constituent rate in Experiment 1. Furthermore, humans and ChatGPT showed similar language-dependent preference to the node-category and parent-category rules (Fig. 3b, $p < 0.001$ for all comparisons, paired two-sided bootstrap, FDR corrected). We also trained an LSTM model to provide a strong baseline for Experiment 2. With only one demonstration, the constituent rate of LSTM was not significantly higher than chance (Fig. 3c, $p = 0.124$ for English and $p = 0.849$ for Chinese, one-sided Monte Carlo test, FDR corrected). The constituent rate of LSTM reached the constituent rate of humans and ChatGPT after about 10 demonstrations for English, but never reached the constituent rate of humans after 700 demonstrations for Chinese. Experiment 2 was also tested on four other LLMs, including GPT-4, Claude-3, Gemini-1, and Llama-3, and the results were generally consistent with the results of ChatGPT (Fig. S4).

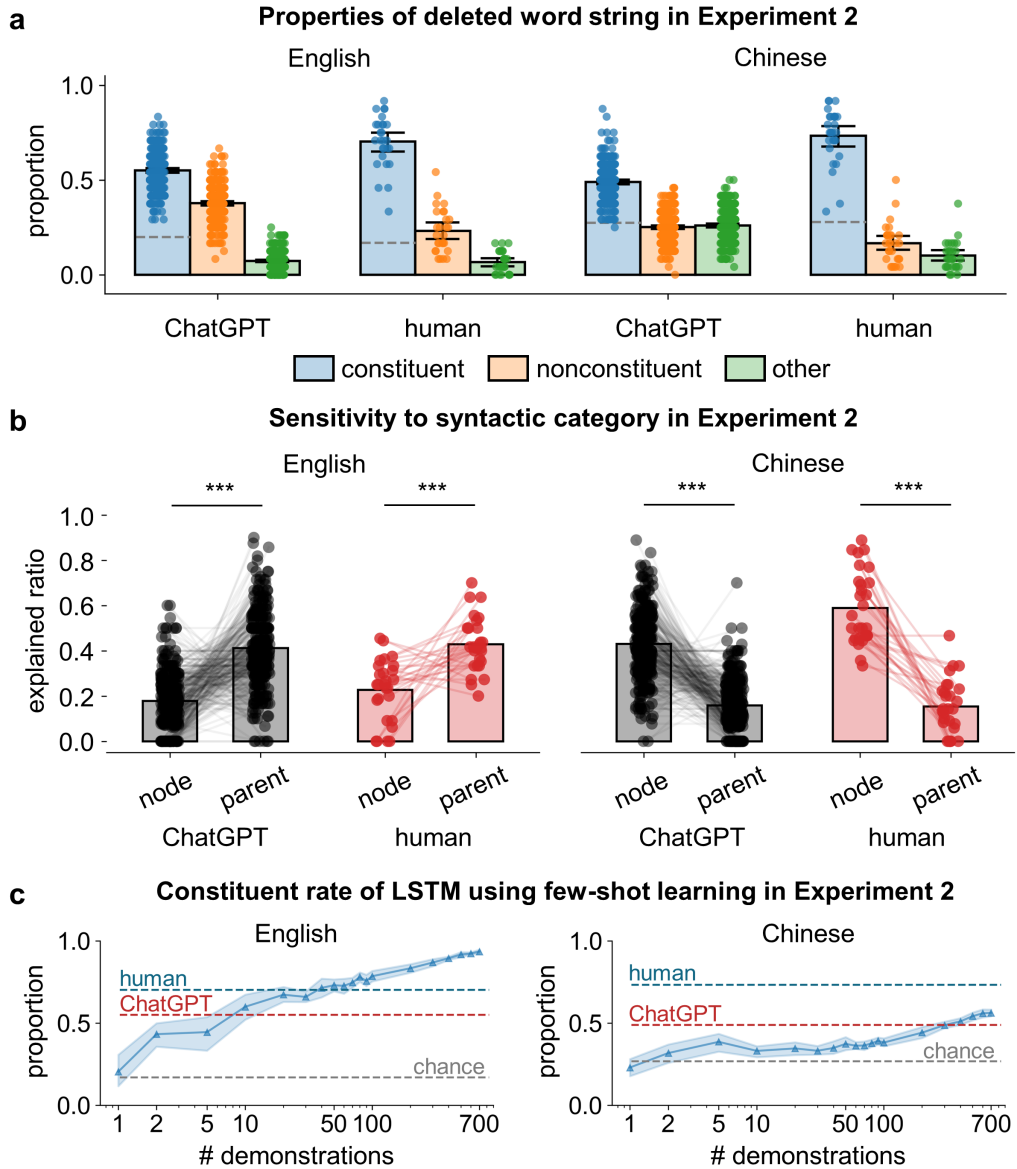

Figure 3. Results of Experiment 2, which deletes an arbitrary constituent, instead of an NP embedded in a VP. **a.** Properties of the deleted word string. Constituent rate is significantly higher than chance. **b.** Explained ratios for the node- and parent-category rules. Participants prefer the node- and parent-category rule for Chinese and English, respectively. **c.** Constituent rate of a naive LSTM that has access to word properties and word position, given different numbers of demonstrations. ****p* < 0.001.

**Constituency Tree Reconstruction and Analysis**

Experiments 1 and 2 demonstrated that human participants and ChatGPT tended to delete a linguistic constituent from the test sentence and the constituent they deleted was influenced by the syntactic category of the constituent deleted in the demonstration. These results highlighted the sensitivity to linguistic constituents and, in Experiment 3, we investigated whether these results could support the reconstruction of the full hierarchical constituency representation of a sentence. We also analyzed whether the constituency tree reconstructed based on human or ChatGPT behaviors matched the linguistic constituency tree. The constituency trees were reconstructed based on the word strings participants deleted from a test sentence when the test sentence was paired with different demonstrations. Specifically, in each test, the participant deleted a word string by analyzing the properties of the word string deleted in the demonstration, and therefore different demonstrations could bias the participants towards deleting different word strings from the same test sentence (Fig. 4a). Consequently, after giving a large number of diverse demonstrations, which deleted different types of constituents (see Methods for details), we could obtain a large variety of deleted strings, which we assumed to correspond to nodes in the latent constituency tree. To characterize which constituency tree best explained the deleted strings, we defined the explained ratio for each potential constituency tree, i.e., the number of deleted strings that were nodes in the tree divided by the total number of deleted strings.

# Experiment 3: Reconstruction of Constituency Tree via Deletion

**a**                        **Procedures for tree reconstruction**
(sentence to analyze: *John found the cat*)

1. Use the sentence as the test sentence, and pair it with different demonstrations to create multiple tests

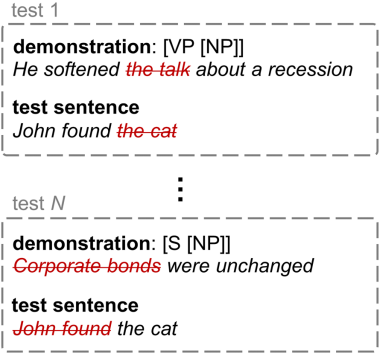

2. A collection of deleted word strings

| string | proportion |
|---|---|
| *the cat* | 0.53 |
| *found the cat* | 0.44 |
| *John found* | 0.03 |

3. Tree reconstruction using CKY algorithm

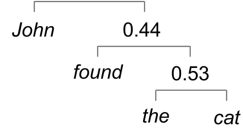

Explained ratio = 0.44 + 0.53 = 0.97
Unexplained strings: *John found* (0.03)

**b**        **Explain deleted word strings using different trees**

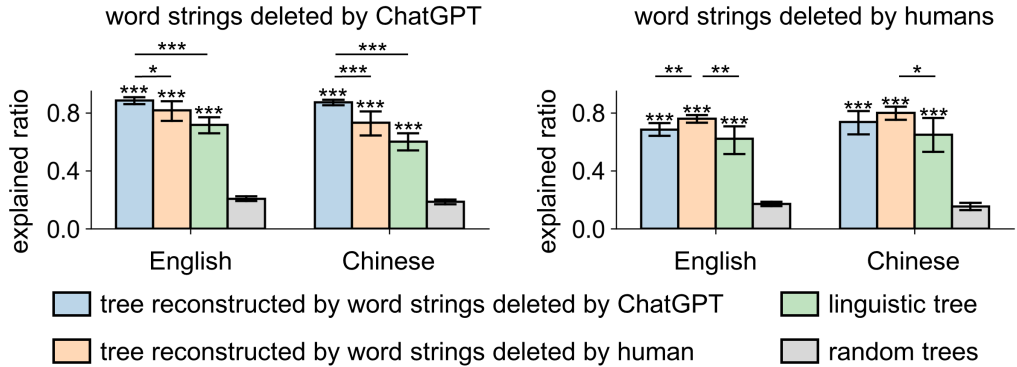

**c**   **F1 score with linguistic trees**       **d**      **Balance factor**

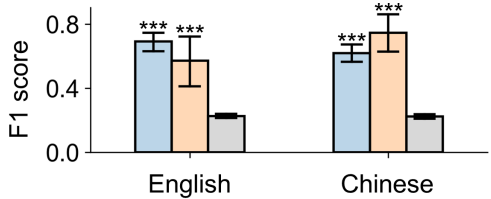 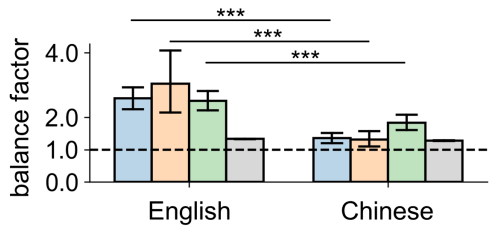

**e**                      **Averaged constituency tree**

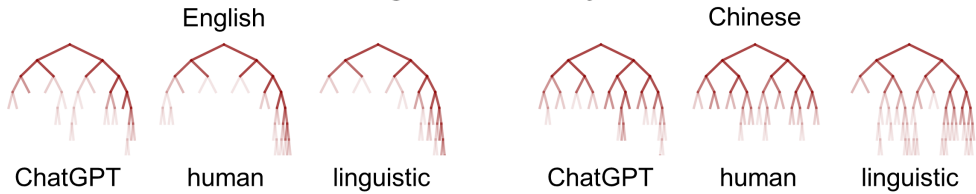

Figure 4. Reconstruction of constituency tree based on word deletion behavior. **a.** Procedures to reconstruct a constituency tree. A test sentence is paired with many demonstrations to generate a variety of deleted word strings, and a word string is explained by a constituency tree if it matches a node in the tree. The proportion of word strings explained by the constituency

tree is called the explained ratio. **b.** Explained ratios for different trees. The linguistic tree is the syntactic tree labeled in the treebank. Comparison with the chance level is shown on top of each bar. **c.** F1 score with respect to the linguistic tree, i.e., the proportion of nodes shared with the linguistic tree. **d.** Balance factor of each constituency tree, i.e., the ratio between the numbers of right descendant nodes and left descendant nodes. Comparisons are only applied between Chinese and English. **e.** Visualization of the constituency trees averaged over all test sentences. *$p < 0.05$, **$p < 0.01$, ***$p < 0.001$.

For example, consider the test sentence "*John found the cat*" and suppose that the participants deleted "*John found*" 30 times, "*found the cat*" 440 times and "*the cat*" 530 times in 1000 tests, each of which was paired with a distinct demonstration (Fig. 4a). For the potential constituency tree [*John* [*found* [*the cat*]]], only "*found the cat*" and "*the cat*" were constituents, and "*John found*" was not. Therefore, the explained ratio of the tree was 0.97, i.e., the total number of tests in which "*found the cat*" or "*the cat*" was deleted divided by the total number of tests. In contrast, for an alternative constituency tree [[*John found*] [*the cat*]], only "*John found*" and "*the cat*" were constituents so that the explained ratio of the tree was 0.56. Consequently, [*John* [*found* [*the cat*]]] better explained the word deletion behaviors than [[*John found*] [*the cat*]]. For a given set of deleted word strings, we used the CKY algorithm to find the binary constituency tree with the highest explained ratio (see Methods). This constituency tree was referred to as the deletion-based constituency tree.

For both humans and ChatGPT, the deletion-based constituency tree explained about 80% of the deleted word strings (blue bars in the left panel of Fig. 4b and orange bars in the right panel of Fig. 4b), significantly higher than chance ($p < 0.001$ for all comparisons, one-sided Monte Carlo test, FDR corrected). To evaluate the similarity between the deletion-based constituency trees and the linguistic constituency tree, we analyzed the overlap in constituents between two trees using the F1 score (see Methods), which was also significantly higher than chance (Fig. 4c, $p < 0.001$ for all comparisons, one-sided Monte Carlo test, FDR corrected).

To further examine the overall characteristics of the deletion-based trees, we averaged the deletion-based trees, as well as the linguistic trees, over all test sentences (Fig. 4e). In general, the deletion-based constituency trees had similar depth and width with the linguistic constituency tree for each language. Furthermore, the linguistic tree tended to have more right descendant nodes in English than Chinese, consistent with the idea that English is more of a right branching language than Chinese (Levy & Manning, 2003; Zhang et al., 2022), and a similar trend was observed for the deletion-based tree. We quantified this trend using balance factors, i.e., the total number of right descendant nodes divided by the total number of left descendant nodes (see Methods). The balance factor was significantly higher in English than Chinese for both the linguistic constituency tree (Fig. 4d, $p < 0.001$, unpaired two-sided bootstrap, FDR corrected) and the deletion-based constituency tree ($p < 0.001$ for both humans and ChatGPT, unpaired two-sided bootstrap, FDR corrected).

These results demonstrated that word deletion behaviors of human and ChatGPT could be well explained by a latent constituency tree, and the latent constituency tree was quantitatively similar to the linguistic constituency tree.

**Semantic Constraints on Constituency Analysis**

In Experiment 3, we reconstructed the constituency trees that could explain word deletion behavior, and demonstrated structural similarities between deletion-based tree and the linguistically defined syntactic tree. Parsing a sentence into structurally interconnected constituents, however, is subject to not only syntactic constraints, but also semantic ones, especially for sentences with ambiguous syntactic structures. In the following Experiment 4, we analyzed whether humans and ChatGPT considered semantic constraints when analyzing syntactically ambiguous constituents. We tested two types of constructions involving structural ambiguity, both of which could be syntactically parsed into two potential constituency trees, referred to as structure 1 and structure 2 (Fig. 5ab). By manipulating the words in the sentence, we constrained that only one structure was associated with a plausible meaning.

# Experiment 4: Semantic Constraints on Constituency

**a**    **Adjunct attachment sentences**

**demonstration:** *NP1 of ~~NP2~~ VP*

*The song of ~~the player~~ is really popular*

**test sentence:** *NP1 of NP2 C VP*

**structure 1**

*The letter of [the writer that had blonde hair] arrived*
                plausible constituent

**structure 2**

*The writer of [the letter that had blonde hair] arrived*
                implausible constituent

☐ target string

**b**    **PP attachment sentences**

**demonstration:** *NP1 verb ~~NP2~~*

*The programmer wrote ~~the code~~*

**test sentence:** *NP1 verb NP2 PP*

**structure 1**

*The guy caught [the rat with a scar]*
              plausible constituent

**structure 2**

*The guy caught [the rat with a trap]*
              implausible constituent

☐ target string

**c**    **Deletion of target string**

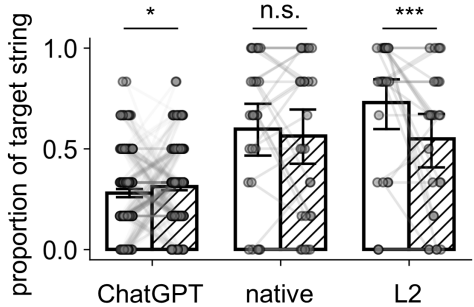

**d**    **Deletion of target string**

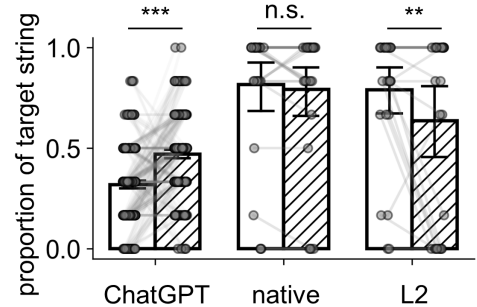

Figure 5. Influence of semantic plausibility on the word deletion task. The test sentence is always a syntactically ambiguous sentence, and only one of the syntactic structures corresponds to a semantically plausible sentence. **a.** In the first type of syntactically ambiguous sentence, the clause *C* (green) could either modify the *NP1* (purple) or *NP2* (blue), forming two potential syntactic structures. The words are selected so that either structure 1 or structure 2 is semantically plausible. The demonstration induces the participants to delete the target string (boxed by the gray line), which is only a semantically plausible constituent in structure 1. **b.** In the second type of syntactically ambiguous sentence, the *PP* (green) could either modify *NP2* (blue) or the verb (purple). **cd.** The proportion of tests in which the target string is deleted for the two types of syntactically ambiguous sentences. The target string is more frequently deleted when structure 1 is semantically plausible for L2 speakers, but not for native speakers and ChatGPT. *$p < 0.05$, **$p < 0.01$, ***$p < 0.001$.

The analysis focused on what we called the target string, i.e., a word string that was a semantically plausible constituent in structure 1 but not structure 2 (see Fig. 5ab). If a participant deleted the target string more frequently when structure 1 was associated with a plausible meaning, i.e., when the target string was a semantically plausible constituent, it suggested that the participant integrated semantic information to determine which constituent to delete. Nevertheless, the results showed that ChatGPT deleted the target string slightly but significantly more frequently when structure 2 was associated with plausible meaning (Fig. 5cd, $p = 0.027$ for adjunct attachment and $p < 0.001$ for PP attachment, paired two-sided bootstrap, FDR corrected), while the frequency of native English speakers deleting the target string was not significantly modulated by the plausibility of structure 1 or 2 ($p = 0.468$ for adjunct attachment and $p = 0.145$ for PP attachment, paired two-sided bootstrap, FDR corrected).

We additionally tested high-proficiency L2 English speakers, and found that L2 speakers deleted the target string more frequently when structure 1 was associated with a plausible meaning ($p < 0.001$ for adjunct attachment and $p = 0.008$ for PP attachment, paired two-sided bootstrap, FDR corrected). These results indicated that humans and ChatGPT mainly relied on syntactic structure instead of semantic plausibility when performing the word deletion task, but high-proficiency non-native speakers were significantly, although not strongly, influenced by semantic plausibility. To further test whether ChatGPT could rely on syntactic structure alone to perform the word deletion task, we additionally constructed syntactically correct nonsense sentences (see Fig. S5a). For these sentences, ChatGPT still deleted constituents much more frequently than chance (Fig. S5b), confirming primary reliance on syntactic structure.

**Discussion**

Using a series of experiments, we demonstrate that humans and LLMs are both sensitive to linguistic constituency when inferring rules in an unfamiliar language task. We could reconstruct a tree-structured constituency representation based on the behavior of human participants and LLMs, and the reconstructed constituency tree is structurally consistent with the linguistically defined constituency tree. These results strongly support the hypothesis that both human participants and LLMs can develop latent representations of linguistic constituents and actively use these representations during language processing.

Although our results strongly support that a latent representation of hierarchical linguistic constituents can emerge in both the human brain and LLMs, it is apparent that humans and LLMs build the representations using different hardware architectures and acquire such representations in different

ways – LLMs learn the representation through a large amount of training text, while humans learn it through a much smaller amount of multimodal data. What we show here is that, after the human brain and LLMs have acquired the basic use of human language (even in different manners), a latent representation of hierarchical linguistic constituents emerges and can functionally explain the behavior during an unfamiliar language task. The results also add to the literature showing that the human brain and LLM, albeit fundamentally different in terms of the implementation, can have aligned internal representations of language (Schrimpf et al., 2021; Goldstein et al., 2022; Stanojević et al., 2023; Lyu et al., 2024; Yu et al., 2024).

**Relation to linguistic constituency tests**

Formal linguistics work defines linguistic constituents based on constituency tests (Carnie, 2002; Adger, 2003). For example, with a substitution test, one can delete a constituent from a sentence and fill in its position with a pro-form (Radford, 1997; Tallerman, 2005). In the current word deletion task, however, the deleted constituent is not filled in and in general the remaining words no longer form a well-formed sentence. The ill-formedness of the remaining words, however, is a key feature of the word deletion task – LLMs are trained using massive text corpora to produce well-formed sentences (Brown et al., 2020), and asking LLMs to generate an improper sentence can potentially alleviate the direct influence of the training data. In other words, the word deletion task creates an out-of-distribution test for LLMs (Bender et al., 2021; Binz & Schulz, 2023; Wilson et al., 2023). Furthermore, classic constituency tests are commonly used by linguists but seldom applied to test naive participants, since they often involve linguistic terminology that is unfamiliar to naive participants. The word deletion task, however, and can be easily described with just one demonstration and does not involve any linguistic terminology. Therefore, the test can be widely used to test tacit constituency knowledge in different LLMs and human populations, including children and clinical populations.

**Syntactic capacity of LLMs**

Previous studies have also probed whether LLMs can correctly parse constituency structures using a number of tasks, such as grammatical judgement or subject-verb agreement (Marvin & Linzen, 2018; Hu et al., 2020; Finlayson et al., 2021). It is well known that, with explicit training, even DNN models much smaller than the current LLMs, such as the LSTM, can achieve high accuracy on syntactic tasks such as constituency parsing (Choe & Charniak, 2016; Cross & Huang, 2016; Kitaev & Klein, 2018) and subject-verb agreement (Linzen et al., 2016; Kuncoro et al., 2018; Linzen & Leonard, 2018). In the current study, with about 50 demonstrations, the LSTM model can also be trained to delete linguistic constituents but fails when only a few demonstrations are provided (Fig. 1d and Fig. 3c). In other words, many DNN models have the computational capacity to perform constituency analysis, at least for relatively simple sentences (Linzen et al., 2016), but whether the models automatically parse the constituency structure without any explicit demand remains much more controversial.

Since human babies, as well as most LLMs, are not explicitly trained on syntactic tasks during language acquisition, many studies focus more on whether syntactic ability can emerge in models not directly trained on syntactic tasks (Michaelov et al., 2023; He et al., 2024). These studies have found that LLMs have similar grammaticality judgement with humans (Golan et al., 2023; Hu et al., 2024). The massive training corpus of LLMs, however, makes it a great challenge to distinguish whether the LLMs have memorized specific examples, learned shallow statistical cues, or truly learned syntactic knowledge (Bender & Koller, 2020; Bender et al., 2021; Mahowald et al., 2023; van Dijk et al., 2023). One solution to the challenge is to test structurally complex sentences that have relatively low occurrence frequency in the training data, e.g., center-embedded sentences (Lakretz et al., 2022; Srivastava et al., 2023). These structurally complex sentences, however, are also challenging to interpret for humans and it is of great interest to also test simpler sentences that humans and LLMs are used to processing. The current word deletion task transforms structurally simple sentences to ill-formed sentences and therefore can be employed to analyze both simple and complex sentences.

Here, we demonstrate that human participants and LLMs are sensitive to linguistic constituency through behavioral analysis. Many previous studies have also demonstrated that an internal syntactic representation emerges in language models by directly analyzing the internal representation of models, i.e., activation of hidden-layer neurons, during language processing (Hewitt & Manning, 2019; Kim et al., 2019; Arps et al., 2022). For example, it has been shown that the internal representation of models falls into different clusters in the vectorial space for word chunks of different syntactic categories, e.g., noun phrases and verb phrases (Jawahar et al., 2019). More importantly, based on the vectorial representations of individual words of a sentence, the tree-structured dependency relationship between words can be decoded (Hewitt & Manning, 2019). Different from the current deletion-behavior-based analysis, these methods all require access to internal activation of language models, which is not possible for closed-source LLMs such as ChatGPT, and cannot directly inform how the decoded constituency representation may contribute to language behavior. When possible, the analysis of internal representation is highly complementary to the analysis of word deletion behavior, since the former can be achieved during arbitrary language processing task while the later can probe whether a constituency representation is actively used during language processing.

**Language-dependent rule inference**

We observe that both human participants and ChatGPT prefer different rules to determine which constituent should be deleted when processing different languages, i.e., Chinese and English. A likely reason is that English is more of a right-branching language, in which the syntactic tree tends to have more right descendent nodes (Levy & Manning, 2003; Zhang et al., 2022). It has been proposed that top-down parsing strategies, which first consider parent nodes, are more memory efficient than bottom-up parsing strategies, which first consider child nodes, for right-branching languages, and the opposite is true for left-branching languages (Mazuka & Lust, 1990; Resnik, 1992). Therefore, it is possible that both humans and LLMs analyze different languages through different constituency parsing strategies, and such strategies are deployed for the word deletion task, resulting in the language-dependent rule inference. Importantly, a recent fMRI study has demonstrated that the neural responses recorded during Chinese and English sentence processing tasks separately fit the predictions of bottom-up and top-down parsers (Zhang et al., 2022). The idea that a person or model can utilize different parsing strategies for different languages is consistent with the principles and parameters framework in linguistics (Chomsky, 1993; Chomsky & Lasnik, 1993) – the structure of different languages shares a number of common principles but differs in its parameters, e.g., head-initial or head-final.

Our data and previous results (Zhang et al., 2022) suggest that humans and LLMs may flexibly switch their analysis parameters based on the input language.

**Influence of syntax and semantics on the word deletion task**

We observed that human participants who are native speakers of the testing language rely more on syntax than semantic plausibility when performing the word deletion task (Experiment 4). A plausible explanation is that the word deletion task does not explicitly require semantic comprehension so that the participants do not actively resolve the two ambiguous interpretations (Townsend & Bever, 2001; Sanford & Sturt, 2002; Ferreira, 2003; Ferreira & Patson, 2007). In contrast, human participants who are proficient but not native speakers of the testing language show evidence of integrating semantic information in the word deletion task, consistent with the idea that L2 speakers focus more on lexico-semantic cues and underuse syntactic analysis (Clahsen & Felser, 2006). Different from the behavior of human participants, ChatGPT shows a weak but significant influence of semantics but tended to delete a semantically implausible constituent. This behavior potentially reflects incorrect integration of semantic and syntactic information, which also echoes previous demonstrations that LLMs trained only with text data often show deficits in world knowledge (Mooney, 2008; Bisk et al., 2020; Marcus, 2020).

In summary, we propose a novel word deletion test that can investigate whether humans and LLMs actively use a latent constituency representation when learning an unfamiliar language task. This test does not rely on any explicit knowledge about linguistic constituents, and can reconstruct a constituency tree based on word deletion behavior. Using this test, we demonstrate that a latent constituency representation emerges in both the human brain and LLMs, and is actively used to solve language processing tasks.

## Methods

### Human participants

The study recruited a total of 338 participants, including 124 native English speakers (18-40 years old, mean 31 years old; 80 females), 124 native Chinese speakers (17-38 years old, mean 23 years old; 70 females), and additionally, 90 native Chinese who spoke English as a second language (18-32 years old, mean 23 years old; 57 females). These bilingual participants all passed the College English Test-6 (CET-6), a national English test system in China, and their average CET-6 score was 538 (roughly equivalent to 6.0 on IELTS). Each experiment included 30 participants with the same language background (except that Experiment 3 included 34 participants). No participants participated in more than one experiment. All participants were provided informed consent before experiments and were financially compensated.

**Code and LLMs**

All scripts were written in Python v3.9.12, using the following packages: pandas v1.4.2 (McKinney, 2010), Numpy v1.22.4 (Harris et al., 2020), SciPy v1.7.3 (Virtanen et al., 2020), Torch v2.2.1 (Paszke et al., 2019), HanLP v2.1.0 (He & Choi, 2021), spaCy v3.4.0 (Honnibal et al., 2020), statsmodels v0.14.1 (Seabold & Perktold, 2010), seaborn v0.12.0 (Waskom, 2021), Matplotlib v3.5.1 (Hunter, 2007), and OpenAI v0.27.2. The online experiments involving human participants were written with Streamlit v1.13.0. All code, sentences and experiment results are available at github.com/y1ny/WordDeletion.

We queried ChatGPT through the OpenAI API (Azure cloud service). All experiments with ChatGPT were conducted using the 'gpt-35-turbo-0301' model variant without any update to model parameters. Following previous studies (Webb et al., 2023), we set the temperature parameter, which controlled the randomness of model response, to 0 and max_tokens, which controlled the maximum number of tokens to generate, to 200 in all experiments. All other parameters were set to their default values. GPT-4, Claude-3, Gemini-1, and Llama-3 were queried through the OpenRouter API, and the experiments with these LLMs were conducted using the 'gpt-4', 'claude-3-sonnet', 'gemini-1-pro', and 'llama-3-70b-instruct' model variants. All parameters were consistent to ChatGPT.

**Metrics to characterize constituency structure**

We summarize metrics to quantify constituency tree in this section. In the constituency tree, the leaf nodes were words in a sentence, which were grouped into latent nodes corresponding to larger linguistic constituents, e.g., phrases. If two nodes, i.e., node A and node B, were directly connected and node A was closer to the terminal nodes, node A was referred to as the *direct child* of node B, and node B was referred to as the *parent* of node A. For example, for [S [NP] [VP]], S was the parent for NP and VP, and NP was the child of S. A *descendant* node could be a direct child, a direct child of a direct child, or a direct child of a direct child of a direct child etc. The *right* child of a node was the child on the right branch (e.g., VP was the right child of S in the structure of [S [NP] [VP]]), and the *right* descendants of a node included its right child and all the descendant nodes of its right child. A node without a parent node was referred to as the root. *Syntactic depth* was the number of edges from the root to the farthest leaf. Each node in the constituency tree had a syntactic category, which was referred to the *node category*. The syntactic category of the parent of a node was referred to as the *parent category*.

**Sentences**

**Treebank sentences.** Treebank sentences used in our study were sourced from the Penn Treebank (PTB) (Marcus et al., 1993) and Chinese Treebank (CTB) (Xue et al., 2005), which included natural sentences and hand-labelled linguistic constituency trees. In Experiment 3, the constituency trees were binarized to compare with the deletion-based trees, following the binarization scheme described in Matsuzaki et al., 2005. We excluded sentences containing named entities, numbers, and punctuations within a sentence, as well as sentences with a 'none' node in their constituency trees. Furthermore, we only kept sentence that had 4 to 15 English words or Chinese characters and had a syntactic depth between 3 to 8. Native speakers were recruited to proofread the sentences and only grammatical sentences were kept. The remaining sentences were referred to as treebank sentences, with the statistics summarized in Table 1.

Table 1. Statistics of treebank sentences used in the study.

| treebank | # sentence | length (words) | | syntactic depth (edges) | |
| :---: | :---: | :---: | :---: | :---: | :---: |
| | | mean | SD | mean | SD |
| PTB | 1251 | 8.6 | 3.0 | 5.6 | 1.4 |
| CTB | 1175 | 8.2 | 2.7 | 5.7 | 1.5 |

**Parallel sentences**. We constructed 60 Chinese sentences and 60 English sentences which had the same syntactic structure and word order (see SI for the full list of sentences). The length of sentences was limited to 5 to 19 English words or Chinese characters. For each language, half of the sentences had the syntactic structure of [S [NP] [VP [NP]]] and the other half had the structure of [S [NP] [VP [PP [NP]]]].

**Syntactically ambiguous sentences**. We collected 12 pairs of English sentences with ambiguous adjunct attachment (Spivey-Knowlton & Sedivy, 1995; Traxler, 2009). The sentences had two noun phrases, i.e., NP1 and NP2, and a relative clause, i.e., C, which were sequenced as "NP1 of NP2 C" and were followed by a VP to form a sentence (e.g., "*The letter of the writer that had blonde hair arrived this morning*"). For each pair of sentences, the noun phrases and the relative clause were the same, but the order of the two noun phrases were switched (see the two sentences in Fig. 5a), so that C should be attached to the either NP2 (structure 1 in Fig. 5a) or NP1 (structure 2 in Fig. 5a) to form a semantically plausible sentence. "NP2 C" formed a plausible constituent for structure 1 (e.g., "*the writer that had blonde hair*") but not structure 2 (e.g., "*the letter that had blonde hair*"), and was referred to as the target string.

We also constructed 18 pairs of English sentences with ambiguous prepositional phrase (PP) attachment (Collins & Brooks, 1995; Schütze, 1995). The sentence always had the structure of "NP1 verb NP2 PP" (e.g., "*The guy caught the rat with a scar*"). The PP was structured as "with NP". Each pair of sentences differed in the PP (and also the verb for some pairs), so that the PP should be attached to either NP2 (structure 1 in Fig. 5b) or the verb (structure 2 in Fig. 5b) to form a semantically plausible sentence. 'NP2 PP' formed a plausible constituent for structure 1 (e.g., "*the rat with a scar*") but not structure 2 (e.g., "*the rate with a trap*"), and was referred to as the target string. The length of syntactically ambiguous sentences was limited to 7 to 15 words. The sentences are listed in SI.

**Demonstration sentences for syntactically ambiguous sentences**. For sentences with ambiguous adjunct attachment, we constructed 30 demonstration sentences using the template "NP1 of NP2 VP" (e.g., "*The forest of the animals pleased us*"). For ambiguous PP attachment sentences, we constructed 30 demonstration sentences using the template "NP1 verb NP2" (e.g., "*The programmer wrote the code*"). The length of demonstration sentences was limited to 5 to 9 words. The demonstration sentences are listed in SI.

**Procedures**

For all experiments, a test consisted of a single demonstration and a single test sentence. In the demonstration, a sentence was shown and a word string was deleted (Fig. 1a). The human participants or LLMs were asked to infer how the sentence was transformed based on the single demonstration and apply the same transformation to a different sentence, referred to as the test sentence. Human data were collected online and a participant had to finish one test before receiving the next test. The participants were reminded in each test that the test was independent of previous tests. For LLMs, we cleared the history after each test to ensure that each test was solved independently. The instructions for human participants, as well as the prompts for humans and LLMs, were provided in SI. The experiment was implemented in either Chinese or English and within each experiment all the stimulus materials and instructions/prompts were in the same language. The selection of sentences in the demonstration and test sentences, as well as the rule governing which a word string was deleted, is detailed in the following subsections.

**Experiment 1**. Experiment 1 was further divided into Experiment 1a, which used treebank sentences, and Experiment 1b, which used parallel Chinese and English sentences. The demonstration sentence ($N$ = 807 for English and $N$ = 716 for Chinese) was required to have an NP, which was deleted. The NP was required to have at least 2 words and was a direct child of a VP (see Metrics to characterize constituency structure). The test sentence ($N$ = 942 for English and $N$ = 837 for Chinese) was required to have an NP that and had at least 2 words and was a descendant of a VP (see Metrics to characterize constituency structure). In Experiment 1a, for each participant or each run of ChatGPT, we randomly sampled 24 demonstration sentences and 24 test sentences from treebank sentences with the requirements mentioned above. For 35.3%/33.2% of the sampled test sentences used in the human/LLM experiment, the NP was a descendant instead of a direct child of the VP and these test sentences were used to dissociate the node and parent category rules. In Experiment 1b, for each participant or each run of ChatGPT, we randomly sampled 24 demonstration sentences and 24 test sentences from parallel Chinese and English sentences. The demonstration sentences were required to have the structure of [S [NP] [VP [NP]]] while the test sentences were required to have the structure of [S [NP] [VP [PP [NP]]]].

Experiment 1a was implemented in both Chinese and English. Thirty native Chinese speakers participated in the Chinese version of the experiment. Thirty native English speakers and 30 Chinese-English bilinguals participated in the English version of the experiment. Two Chinese speakers and two English speakers were excluded from the analysis since they gave the same answer to every question ($N$ = 1) or consistently typed in random characters

($N$ = 3). Experiment 1b was also implemented in both Chinese and English. Thirty native Chinese speakers participated in the Chinese version of the experiment, and 30 Chinese-English bilinguals participated in the English version of the experiment. For both the Chinese and English versions of the experiment, ChatGPT received 7200 tests, which were divided into 300 runs and each run had the same number of tests ($N$ = 24) as in the human experiment. The results within a run were averaged as in the analysis of the human experiment.

**Experiment 2.** In Experiment 2, for each participant or each run of LLM, we randomly sampled 24 demonstration sentences and 24 test sentences from treebank sentences with the following procedure. Each demonstration sentence was randomly selected from treebank sentences, and an arbitrary constituent was deleted. For each demonstration, the corresponding test sentence was randomly selected from treebank sentences with the following constraints: First, the sentence must contain a constituent that had the same node category as the deleted constituent in the demonstration (see Metrics to characterize constituency structure). Second, the same constituent or a different constituent in the sentence must have the same parent category as the deleted constituent in the demonstration (see Metrics to characterize constituency structure). As a consequence of this random sampling procedure, for 23.9%/28.1% of the test sentences used in the human/LLMs experiment, one constituent had the same node category as the deleted constituent while another constituent had the same parent category, and these tests were used to dissociate the node and parent category rules.

Experiment 2 was implemented in both Chinese and English. Thirty native Chinese speakers participated in the Chinese version of the experiment and 30 native English speakers participated in the English version of the experiment. Two Chinese speakers and three English speakers were excluded from the analysis since they gave the same answer to every question ($N = 2$) or consistently typed in random characters ($N = 3$). For both the Chinese and English versions of the experiment, ChatGPT received 7200 tests, which were divided into 300 runs and each run had the same number of tests ($N = 24$) as in the human experiment. Four other LLMs were tested and each received 720 tests.

**Experiment 3.** In Experiment 3, we reconstructed the full constituency tree for a few test sentences ($N = 60$ for ChatGPT and $N = 12$ for humans) based on the word deletion task. Fewer sentences were tested for humans since the experiment was time-consuming. The 60 test sentences for ChatGPT were randomly selected from treebank sentences, with the constraint that the syntactic depth was between 3 and 8 (10 sentences selected for each syntactic depth). The experiment was also presented both in Chinese and

English, and 60 sentences were selected for each language. The 12 test sentences for humans were randomly selected from the 60 sentences, with the constraint that 2 sentences were selected for each syntactic depth.

Each test sentence was paired with multiple demonstration sentences. To increase the diversity of the deleted constituent in the demonstration, we considered all possible combinations of node and parent categories that existed in the treebank. For ChatGPT, for each possible combination of node and parent categories (e.g., node category = NP and parent category = VP), we randomly selected 10 constituents to construct the demonstration (fewer constituents were selected if fewer than 10 such constituents existed in the treebank). These constituents were deleted from the corresponding sentences in the demonstration. In total, 141/56 combinations of node and parent categories were identified for Chinese/English, and 833/354 demonstration sentences were selected. For human participants, to control the experiment time, we only retained the 50 most common combinations of node and parent categories and selected 2 demonstration sentences for each combination (i.e., 100 tests for each of the 12 sentences).

Each human participant received 36 tests, which included all 12 test sentences, each of which was paired with 3 randomly selected demonstration sentences. The 36 tests were presented in a randomized order. To cover all

100 × 12 tests for humans, we recruited 34 native participants for the Chinese and English version of the experiment. Four Chinese speakers and six English speakers were excluded from the analysis since they gave the same answer to every question ($N = 3$) or consistently typed in random characters ($N = 7$). Results from all human participants (or ChatGPT runs) were pooled to reconstruct the constituency tree of each test sentence.

**Experiment 4.** In Experiment 4, for each participant or each run of ChatGPT, we designed a balanced test construction based on the syntactically ambiguous sentences: a total of 24 tests consisting of 6 adjunct-1 sentences (i.e., adjunct attachment sentences where structure 1 is associated with a plausible meaning), 6 adjunct-2 sentences, 6 PP-1 sentences, and 6 PP-2 sentences. Within these 24 tests, only one sentence from each pair of syntactically ambiguous sentences could appear. Each ambiguous sentence was randomly paired with the corresponding demonstration sentences (see Sentences). For both demonstration sentences of adjunct attachment and PP attachment, the NP2 was deleted.

Experiment 4 was implemented in English. Thirty native English speakers and 30 Chinese-English bilinguals participated in the experiment. One native speaker and three bilinguals were excluded from the analysis since they gave the same answer to every question ($N = 2$) or consistently typed in random characters (*N* = 2). ChatGPT received 7200 tests, which were divided into 300 runs and each run had the same construction and the same number of tests (*N* = 24) as in the human experiment.

**LSTM Modeling**

We employed the Long Short-Term Memory network (LSTM) (Hochreiter & Schmidhuber, 1997) as a baseline model to test whether word deletion task could be learned based on statistical cues derived based on a linear word sequence. These statistical cues included word properties (captured by word vectors) and their ordinal positions in the sentences. The LSTM was a gated recurrent neural network and here we employed a naive LSTM that was not pretrained on human language and therefore did not have a preexisting mechanism to analyze linguistic constituents. The LSTM results under one-shot learning were treated as a baseline and we also trained LSTM with multiple demonstrations to probe how many samples were enough to learn the word deletion task based on these linear-sequence statistical cues.

The input to the LSTM was a complete sentence and each English word or Chinese character was represented by a pretrained vector: The pretrained vector for English words was GloVe (Pennington et al., 2014) and the pretrained vector for Chinese characters was SGNS (Li et al., 2018). To explicitly encode the ordinal position of each word, sinusoidal position embeddings were added to each word or character vector (Vaswani et al., 2017). The output of the LSTM consisted of two numbers corresponding to the positions of the first and last words of the word string to delete. The LSTM included an encoder and a pointer network (Vinyals et al., 2015). The encoder processed the input sentence and generated a contextualized vector for each word, which was fed into the pointer net to identify the word string to delete. The LSTM modeling was employed in Experiments 1 and 2. We varied the size of the training set from 1 to 700. The number of training steps was always 500. The trained LSTM was evaluated on 100 test sentences. We trained and tested the LSTM 10 times for each condition to obtain the final results. All experiments used an adamax optimizer (Kingma & Ba, 2015), a learning rate of 0.002, and a batch size of 32. The hyper-parameters of the LSTM's architecture and training followed the setting of match-LSTM (Wang & Jiang, 2017).

## Data analysis

**Classification of deleted string and constituent rate.** The word string a human participant or LLM deleted in each test was classified into one of three classes, i.e., (1) constituents, i.e., the deleted string was a complete constituent, (2) nonconstituents, i.e., the deleted strings was not a complete constituent, (3) others, i.e., invalid responses in which the participants added in words not in the original sentence or did not delete any word. The constituent rate was the proportion of deleted word strings falling into the 'constituents' category and was calculated for each human participant or each run of LLM.

**Explained ratio.** In Experiment 1 and 2, we employed the explained ratio to quantify how the deleted word strings could be explained by a rule, i.e., the node-/parent-category rule. For instance, an explained ratio of 0.6 for the node-category rule indicated that, for 60% of the deleted word strings, the node category of deleted constituents in the test sentence was consistent to that in the demonstration. The explained ratio was determined for each participant or each run of LLM, and was only calculated based on the tests in which a constituent was deleted.

In Experiment 3, the explained ratio also was employed to quantify how the deleted word strings could be explained by a constituency tree. In this case, it was defined as the number of deleted strings that were nodes in the tree divided by the total number of deleted strings. The explained ratio of a constituency tree was calculated for each test sentence based on all tests pooled over participants or runs of LLM.

**Reconstruction of constituency tree.** In Experiment 3, we reconstructed the constituency tree based on the word deletion task using the Cocke-Kasami-

Younger (CKY) algorithm (Kasami, 1966; Younger, 1967). In this experiment, each test sentence was paired with multiple demonstrations to form multiple tests, and the word strings deleted in all these tests were pooled to reconstruct a constituency tree of the test sentence. We calculated the frequency of each deleted word string in the pool and applied the CKY algorithm to generate a binary constituency tree that had highest explained ratio.

**F1 score**. The F1 score measured the overlap in constituents between the deletion-based constituency trees and linguistic constituency trees. It was defined as $F1 = 2O / (L + D)$, where $D$ was the total number of constituents in the deletion-based tree, $L$ was the total number of constituents in the linguistic tree, and $O$ was the number of constituents that overlapped between the deletion-based and linguistic trees.

**Balance factor**. The balance factor measured whether a constituency tree was left-branching or right-branching, and was defined as the total number of right descendants across all nodes divided by the total number of left descendants across all nodes. A constituency tree was considered completely balanced when the balance factor equaled 1. The tree was more right-branching or left-branching if the factor was greater or less than 1.

**Statistical analysis**

**Monte Carlo test.** One-sided nonparametric Monte Carlo significance tests (Hope, 1968) were used to test whether a response pattern was significantly above chance. The chance level was estimated based on 1,000 simulations. If the actual response was smaller than *A* out of the 1,000 stimulated samples, the significance level was (*A* + 1)/1001 (one-sided comparison). To calculate the chance level for constituent rate, we randomly deleted word strings, i.e., any subset of words that were connected in the original sentence, from all test sentences (*N* = 720 for human participants, 7,200 for ChatGPT, 720 for other LLMs, and 100 for LSTM), and calculated the constituent rate of these randomly deleted word strings. The procedure was repeated 1,000 times, creating 1,000 chance-level constituent rates.

To calculate the chance level for explained ratio, F1 score, and balance factor, we randomly deleted word strings from the same test sentence the same number of times in the reconstruction experiment, reconstructed the constituency tree based on these randomly deleted word strings, and calculated the corresponding metrics. The procedure was repeated 1,000 times, creating 1,000 chance level constituency trees per test sentence.

**Bootstrap.** Bias-corrected and accelerated bootstrap was used to compare the metrics across conditions (Efron & Tibshirani, 1994). For paired comparisons, the participants (or runs for LLM and LSTM) were resampled 10,000 times with replacement and each time the difference across conditions was averaged across participants (or runs for LLM and LSTM), producing 10,000 resampled mean differences. For unpaired comparisons, the participants (or runs for LLM and LSTM) were separately resampled 10,000 times with replacement for each condition, and each time the difference across conditions was averaged across participants (or runs for LLM and LSTM), producing 10,000 resampled mean differences. For both paired and unpaired comparisons, if $A$ out of the 10,000 resampled differences were greater or smaller than 0 (the smaller value), the significance level was $2(A+1)/10001$ (two-sided comparison). When multiple comparisons were performed, the p-value was further adjusted using the FDR correction. Bootstrap was also used to estimate the CI across participants, and for this purpose participants were resampled 1,000 times with replacement.

**Acknowledgements**

We thank Shaonan Wang for inspiring discussions, Lingjun Jin and Xunyi Pan for helping to construct ambiguous sentences, Max Wolpert and Felix Hao Wang for constructive comments on earlier versions of the manuscript. The study is supported by the National Key Research and Development Program of China (No. 2021ZD0204105).

**Author contributions**

N.D. conceived the study. N.D., M.X., and W.L. designed the experiment. W.L. implemented and conducted the experiments. W.L. analyzed the data. N.D., M.X., and W.L. wrote the manuscript.

**Competing interests**

The authors declare no competing interests.

**Additional information**

Supplementary information: Fig. S1-S5, lists of parallel sentences, syntactically ambiguous sentences, and demonstration sentences for syntactically ambiguous sentences, as well as the instructions and prompts for human participants and LLMs.